%% file: umap2020-sigconf.tex
\PassOptionsToPackage{warn}{textcomp}

\documentclass[sigconf]{acmart}

\usepackage{amsmath}
\usepackage{booktabs} 
\usepackage{times}
\usepackage{latexsym}

\usepackage{url}
\usepackage{times}
\usepackage{latexsym}

\usepackage{url}
\usepackage{xcolor}

\usepackage{enumitem}
\usepackage{verbatim}
\usepackage{booktabs} 
\usepackage{multirow}
\usepackage{amsbsy}
\usepackage{array}
\usepackage{booktabs,makecell,tabularx}

\usepackage{arydshln}
\usepackage{changepage}
\usepackage{siunitx}

\newcolumntype{R}[1]{>{\raggedleft\arraybackslash}p{#1}}
\newcolumntype{L}[1]{>{\raggedright\arraybackslash}p{#1}}
\newcolumntype{C}[1]{>{\centering\let\newline\\\arraybackslash\hspace{0pt}}m{#1}}
\usepackage{relsize}
\let\olditemize\itemize
\let\endolditemize\enditemize
\renewenvironment{itemize}{%
    \smaller
    \olditemize
}{%
    \endolditemize
}

\usepackage{array}
\definecolor{green1}{HTML}{257101}
\definecolor{blue1}{HTML}{2C05A5}
\definecolor{red1}{HTML}{A50520}

\AtBeginDocument{%
  \providecommand\BibTeX{{%
    \normalfont B\kern-0.5em{\scshape i\kern-0.25em b}\kern-0.8em\TeX}}}

\acmConference[In submission]{}{2020}{}
\acmBooktitle{In submission}
\acmPrice{}
\acmISBN{}

\renewcommand\footnotetextcopyrightpermission[1]{} 
\begin{document}

\title{Predicting User Engagement Status for Online Evaluation of Intelligent Assistants}
\subtitle{}

\author{Rui Meng}
\orcid{1234-5678-9012}
\affiliation{%
  \institution{University of Pittsburgh}
  \streetaddress{135 N. Bellefield Ave}
  \city{Pittsburgh} 
  \state{PA} 
  \country{USA}
  \postcode{15260}
}
\email{rui.meng@pitt.edu}

\author{Zhen Yue}
\affiliation{%
  \institution{Disney Streaming Service}
  \streetaddress{}
  \city{Los Angeles} 
  \state{CA} 
  \country{USA}
}
\email{zhen.yue@disney.com}

\author{Alyssa Glass}
\affiliation{%
  \institution{Apple Inc.}
  \streetaddress{}
  \city{Cupertino} 
  \state{CA} 
  \country{USA}
}
\email{aglass.ca@gmail.com}

\renewcommand{\shortauthors}{R. Meng et al.}


\begin{abstract}
Evaluation of intelligent assistants in large-scale and online settings remains an open challenge. User behavior based online evaluation metrics have demonstrated great effectiveness for monitoring large-scale web search and recommender systems. Therefore, we consider predicting user engagement status as the very first and critical step to online evaluation for intelligent assistants. In this work, we first proposed a novel framework for classifying user engagement status into four categories -- fulfillment, continuation, reformulation and abandonment. We then demonstrated how to design simple but indicative metrics based on the framework to quantify user engagement levels. We also aim for automating user engagement prediction with machine learning methods. We compare various models and features for predicting engagement status using four real-world datasets. We conducted detailed analyses on features and failure cases to discuss the performance of current models as well as challenges.

\end{abstract}



\maketitle

\input{umap2020-body.tex}


\end{document}

%% file: umap2020-body.tex
\vspace{-1ex}
\section{Introduction}
The increasing popularity of intelligent assistants such as Alexa, Siri and Google Home has attracted broad attention to human-machine dialogue systems, but also brought challenges for evaluating the performance of dialogue systems in online environments. Previous research demonstrated that the most effective way to improve any online system is to optimize it for end-user engagement~\cite{deng2016data}. For example, recommender systems can be optimized for user click and dwell time~\cite{yi2014beyond} and web search systems can be optimized for click-through rate~\cite{graepel2010web} and reformulation rate~\cite{hassan2013beyond}. 
Researchers on dialogue systems have started developing novel evaluation methods from the user's perspective~\cite{jiang2015automatic, sano2017predicting}, nevertheless designing proper metrics for better optimizing online intelligent assistant systems remains a big challenge.

Previous studies seeking to evaluate dialogues systems mainly focus on individual system component performance rather than overall user engagement. The common practice in system-oriented evaluation is breaking down the dialogue system into parts, such as dialogue act classification and state tracking, and evaluating the performance of each component respectively. However, we cannot assess the performance of the whole system by simply aggregating the performance of each component. There were several methods developed to evaluate the overall system performance. For example, one can evaluate the quality of system responses by measuring their similarities to ground-truth responses with metrics like BLEU~\cite{sordoni2015neural, vinyals2015neural}. However, users' requests in online environments are very diverse and dynamic and may not be covered by the ground-truth based evaluation.
In addition, it is very expensive to build ground-truth datasets to cover a variety of domains and the lack of automatic methods 
makes evaluation hard to scale up for online scenarios. 

Research in web search has a long history of conducting large-scale online evaluation for search engines utilizing user engagement and behavior signals~\cite{diriye2012leaving, hassan2010beyond, hassan2011task, hassan2013beyond}. The idea was to regard possible user interaction outcomes as different engagement types, such as long-dwell click, query reformulation and abandonment. These engagement types can then be used to gauge search success and cost, thus making these measurements scalable for online evaluation. We think that the same idea can be adopted to the evaluation of intelligent assistants as well. For example, we can classify each user utterance in a dialogue system into success and failure requests. Previous research proposed a conceptual framework PARADISE~\cite{walker1997paradise} for evaluating dialogue systems. It pointed out that a successful dialogue system should maximize task success and minimize cost. Same for the online evaluation of intelligent assistant, we should not only focus on whether or not users' requests have been fulfilled but also measure how much effort it takes. We cannot simply use the conversation length or number of turns as measurement for cost, since it might take multiple necessary turns to finish a complex user request. Instead, we should focus on whether or not the interaction is necessary for the intelligent assistant to fulfill the request. In order to solve the problem, we proposed a novel scheme categorizing users' utterances into different engagement status, with which we can design metrics to measure task success and cost for online evaluation of intelligent assistant.

Furthermore, we aim for a more challenging task, delivering an automatic method for predicting the user engagement status. In recommendation and search, researchers utilize behavior signals such as dwell time and query content features to automatically predict user engagement. Similarly, we utilize interaction signals between users and intelligent assistants to predict users' engagement status. Comparing to the short text in queries, user interactions with intelligent assistants have rich content information, which can be used for creating more sophisticated features. We then studied various machine learning methods and feature settings for this prediction task using four annotated datasets. 

The contributions of this paper are summarized as follows: (1) we introduced a novel utterance-level scheme for classifying user engagement status. It takes into account both success and cost to measure to what extent the user is engaged or satisfied\footnote{``User Engagement'' and ``User Satisfaction'' both are common user experience evaluation metrics~\cite{attfield2011towards, dupret2013absence}. ``User Satisfaction'' is usually related to self-reported judgments and can be considered as a specific aspect of the user engagement study~\cite{o2018practical}. We use the more general concept ``User Engagement'' because this study is not limited only to self-reported judgments.} when interacting with an intelligent assistant; (2) we employ both classic machine learning and deep learning techniques to automatically predict user engagement status using various features extracted from dialogues, and provide comprehensive and thorough analyses on four real-world datasets.


\vspace{-1ex}
\section{Related Work}
\subsection{Evaluation of Intelligent Assistants}
There are four major methods being widely used for evaluating intelligent assistants: (1) Evaluation on specific components~\cite{kim2010classifying,ohtake2008unsupervised, griol2016neural}. People have established several tasks to examine certain aspects of the systems, such as dialog state tracking and dialogue act classification, and evaluate them by metrics like precision and recall. While these evaluations are useful to identify problems in each component, the outcomes cannot reflect the overall performance of the dialogue system. (2) Evaluation by comparing system responses with ground-truth responses~\cite{ritter2011data, sordoni2015neural, li2015diversity}. This type of approaches is broadly adopted for response generation. The basic idea is we measure how appropriate a proposed response is by checking its similarity to ground-truth responses with metrics like BLEU~\cite{papineni2002bleu} and METEOR~\cite{banerjee2005meteor}. However, a high degree of token matching may imply its readability, but does not mean it is a logical response, and such methods have been proved correlated poorly with the human judgment~\cite{liu2016not}. (3) Human evaluation based on Mechanic Turk~\cite{li2016deep, lowe2016evaluation, serban2017hierarchical}. A set of studies employed crowdsourced workers to evaluate system performance. Undeniably this is the most reliable and direct way for evaluating any system, but it can only be used in small-scale off-line studies. Whereas for online intelligent assistants, it requires monitoring their real-time performances. (4) There are a few tasks aiming to detect problematic system responses which share a similar motivation to our study, such as error detection~\cite{krahmer2001error, meena2015automatic} and breakdown detection~\cite{higashinaka2016dialogue}. But in these tasks, the cost of communication is not considered and task boundaries are presumably given. In the real world, both task success and cost affect users' experience considerably and users can move to a new task anytime, therefore our specially designed framework, detecting both system failures and user request boundaries, are more suitable for evaluating real-world intelligent assistant systems.

\vspace{-1ex}
\subsection{User Engagement Prediction}

User Engagement is a quality of user experience characterized by the depth of a user's investment when interacting with a digital system~\cite{o2016theoretical}. Studies of both user behavioral metrics (e.g. web page visits and dwell time) and self-reported judgments (e.g. satisfaction) can be considered as specific aspects of user engagement study~\cite{o2018practical}, and our work is also closely related to them. 

User satisfaction rating in dialogue systems has been discussed for a long time~\cite{kamm1995user, shriberg1992human, polifroni1992experiments}. A wide variety of techniques and features has been studies~\cite{yang2010collaborative, chowdhury2016predicting, hara2010estimation}, as well as some recent efforts on the basis of deep neural networks~\cite{lowe2017towards}. Most of these studies output a holistic satisfaction rating for the entire dialogue, but it cannot offer any further information about how the system fails to satisfy users. Therefore it is not a reliable optimization target that can be used for improving the dialogue system. 

PARADISE~\cite{walker1997paradise} framework tackles this problem by breaking down the measurement of user satisfaction into two parts: task success and dialogue cost. However, PARADISE was proposed more than twenty years ago and is more a conceptual framework than a practical solution: the two factors are too general to implement and they did not provide any automatic method practicable in a nowadays large-scale setting. For example, to measure task success it requires conversations to be represented as task-specific attribute-value pairs which are not flexible nor transferable. The measurements of dialogue cost using utterance length and number of dialogue turns are also arguable, because for many user requests, say the restaurant inquiry in DSTC2 (see Table \ref{tab:dstc2_example}), it naturally takes several necessary turns to finish. Taking PARADISE as conceptual inspiration, we greatly modify it to fit the online evaluation requirement of nowaday intelligent assistants, by proposing a novel utterance classification scheme as well as examining a series of automatic methods. These two contributions make our study significantly different from the PARADISE, yielding sound outputs from the perspective of practice.

Our work is also similar to \cite{jiang2015automatic, kiseleva2016predicting} in the sense of using user behavior signals to predict short-term user engagement/satisfaction. They focus on voice search tasks and use simple statistic features borrowed from traditional information retrieval studies, while ignoring the rich text features in the utterances. In contrast, our study utilizes a great variety of language features from the dialogue content, which we think can be essentially better to reveal the real user intents.

\vspace{-1ex}
\section{Classifying and measuring user engagement}
In this section, we introduce a novel classification framework of user engagement status and describe how we utilize this framework to design metrics that can measure both success and cost of dialogues.

\vspace{-1ex}
\subsection{Framework for Classifying User Engagement}
\label{sec:classification_scheme}

Prior to introducing the classification framework, we would like to clarify several concepts that play important roles in it. We consider a continuous interaction between user and system within a small range of time as a \textit{session}. And a \textit{session} is comprised of a number of \textit{utterances}, each of which is issued by either the user or the system. We can further group \textit{utterances} into \textit{tasks} basing upon user's potential information needs. For example, Table \ref{tab:dstc2_example} shows a \textit{session} sampled from the dataset DSTC2, in which a user (\textbf{User}) consults a dialogue system (\textbf{Bot}) about restaurants and the system asks the user to provide necessary information to narrow down the scope. It consists of 7 effective pairs of \textit{user utterance} and \textit{system utterance}, and 3 basic \textit{tasks} (or \textit{user requests}): requesting a restaurant (\#1-\#3), requesting its address (\#4), and requesting its phone number (\#5) and requesting its postcode (\#6-\#7). 

With the goal of measuring both success and cost of user interaction with intelligent assistant, we propose a four-class utterance classification scheme. Each class represents the engagement status of a user after issuing an utterance. 

\begin{itemize}[leftmargin=*,topsep=6pt]
\fontsize{9}{9}\selectfont
\setlength\itemsep{0.5em}
\item \textbf{\textit{Fulfillment} (F)}: current user request is understood and fulfilled by the system.
\item \textbf{\textit{Continuation} (C)}: current user request is understood by the system but more interactions are required to complete this request. 
\item \textbf{\textit{Reformulation} (R)}: current user request is wrongly or incompletely understood by the system, and user repeats/rephrases this request in the next turn.
\item \textbf{\textit{Abandonment} (A)}: current user request is not understood or fulfilled by the system, and the user abandons this request by closing the conversation or starting a new request.
\end{itemize}


\begin{table}[!htbp]
    \centering
    \footnotesize
    \vspace{-2ex}
    \caption{An annotated session from \textbf{\textit{DSTC2}}.}
    \label{tab:dstc2_example}
 	\vspace{-3ex}
    \begin{tabular}{C{0.65cm}|rL{4.0cm}|C{0.65cm}|C{0.8cm}}       
        \toprule\hline
        \textbf{\# Turn} & & \multicolumn{1}{c|}{\textbf{Utterance}} & \textbf{Label} & \textbf{Task} \\\hline
		& \textbf{Bot:} & Hello, welcome to the Cambridge restaurant system? You can ask for restaurants by area, price range or food type. How may I help you? &  &  \\\hline
		1 & \textbf{User:} & Moderately priced & R & \multirow{6}{*}{\textcolor{green1}{\textbf{Success}}} \\\cdashline{2-3}
		& \textbf{Bot:} & What part of town do you have in mind? &  \\\cline{1-4}
		2 & \textbf{User:} & North & C \\\cdashline{2-3}
		& \textbf{Bot:} & Would you like something in the cheap, moderate, or expensive price range? &  \\\cline{1-4}
		3 & \textbf{User:} & Moderate & F \\\cdashline{2-3}
		& \textbf{Bot:} & Golden wok is a nice restaurant in the north of town in the moderate price range &  \\\hline
		4 & \textbf{User:} & Address & F & \multirow{2}{*}{\textcolor{green1}{\textbf{Success}}} \\\cdashline{2-3}
		& \textbf{Bot:} & Sure , golden wok is on 191 Histon Road Chesterton &  \\\hline
		5 & \textbf{User:} & Phone number & F &  \multirow{2}{*}{\textcolor{green1}{\textbf{Success}}} \\\cdashline{2-3}
		& \textbf{Bot:} & The phone number of golden wok is 01223 350688 &  \\\hline
		6 & \textbf{User:} & Postcode & R & \multirow{6}{*}{\textcolor{red1}{\textbf{Failure}}} \\\cdashline{2-3}
		& \textbf{Bot:} & Golden wok is a nice restaurant in the north of town in the moderate price range &  \\\cline{1-4}
		7 & \textbf{User:} & Postcode & A \\\cdashline{2-3}
		& \textbf{Bot:} & Golden wok is a nice restaurant in the north of town in the moderate price range &  \\\hline
		8 & \textbf{User:} & Thank you good bye  &         \\
        \bottomrule
    \end{tabular}
    \vspace{-2ex}
\end{table}

In Table \ref{tab:dstc2_example}, user utterances are annotated with the proposed classification scheme, as shown in the rightmost column. Specifically, the user told the system her desired price range and location (\textit{Turn \#1} and \textit{\#2}), but the system failed to catch the first price information. After the user repeated it (\textit{Turn \#3}), the system returned a restaurant that the user might be interested in. It is worth noting that the annotation of an utterance $utt_i$ has to be one-turn delayed, determined after knowing the future responses ($utt_{i+}$) from both the system and the user side. Therefore the \textit{Turn \#1} utterance is annotated as `\textbf{R}'. The system replied correctly in both \textit{Turn \#4} and \textit{\#5}. The user requested the postcode in \textit{Turn \#6} and repeated it in the \textit{Turn \#7}, and in the end she terminated the conversation after an incorrect response.  Thus \#6 is labeled as `\textbf{R}' and \#7 is `\textbf{A}'.

\begin{table}[!htbp]
    \centering
    \vspace{-2ex}
    \caption{Illustration of two dimensions along which the proposed utterance classification scheme can be binarized.}
    \label{tab:binarized_scheme}
 	\begin{tabular}{r|cc}
		\hline\hline
        & \textbf{Ongoing} & \textbf{Ending} \\ \hline
        \textbf{Correctly Responded} & \multicolumn{1}{l|}{Continuation} & Fulfillment \\ \cline{2-3} 
		\textbf{Wrongly Responded} & \multicolumn{1}{l|}{Reformulation } & Abandonment\\
        \bottomrule
    \end{tabular}
    \vspace{-2ex}
\end{table}

From the definition of each type and the examples, we can see that the proposed classification scheme is clearly defined and highly explainable, because the four classes of user utterance are mutually exclusive and each depicts an explicit user behavior. As shown in Table \ref{tab:binarized_scheme}, our scheme can be thought as two orthogonal binary classifications by checking (1) if the user continues or terminates the current task/request and (2) if the system gives a correct or wrong response. Based on the two conditions, one can assign labels much easier than giving a subjective score~\cite{yang2010collaborative, lowe2017towards} or a sentiment class~\cite{chowdhury2016predicting}. For example, we can split the session in Table \ref{tab:dstc2_example} into four tasks and classify them in into \textcolor{green1}{\textbf{Success}} or \textcolor{red1}{\textbf{Failure}} using \textbf{F} or \textbf{A} as task boundary and satisfaction indicator.

\vspace{-1ex}
\subsection{Online Evaluation Metrics based on User Engagement Status}
\label{sec:metrics}
In the context of industrial web services, ahead of optimizing any system to improve its performance for end users, it is common that we first determine how to measure the user engagement with the system, with the goal of creating engagement metrics that accurately reflect the performance of the product. 
With the proposed classification scheme, not only are we able to understand the engagement status of a user after each request immediately, it also enables us to define a series of evaluation metrics to monitor the system performance in an online manner. Similar to PARADISE~\cite{walker1997paradise}, we would like to define two metrics, from the aspect of success and cost respectively, to measure the user engagement. 

On one hand, since \textit{Fulfillment} or \textit{Abandonment} indicates the boundary of a task as well as a good/bad user experience, we can split a session to several tasks with these two labels, and then group them into successful/unsuccessful tasks respectively. And we define the \textbf{Success Rate} of a session $\mathbb{S}$ as the percentage of success tasks as in Eq. \eqref{eq:success_rate}, where $\#(TASK_{success \in \mathbb{S}})$/$\#(TASK_{ \in \mathbb{S}})$ denotes the number of successful/all tasks in the session $\mathbb{S}$:

\vspace{-1ex}
\begin{equation}
\mathit{SuccessRate}=\frac{\#(TASK_{success \in \mathbb{S}})}{\#(TASK_{ \in \mathbb{S}})}\label{eq:success_rate}
\end{equation}

On the other hand, we would like a metric to represent how efficiently a system can respond to each user request. Firstly, we can use a statistic of \textit{Reformation} to represent the degree to which a user repeats in a task. We define the \textbf{Reformulation Rate} of a session as the percentage of reformulated utterances in each task as in Eq. \eqref{eq:reform_rate}, where $\#(UTT_{reform \in \mathbb{S}})$/$\#(UTT_{\in \mathbb{S}})$ denotes the number of \textit{Reformulation}/all user utterances in the session $\mathbb{S}$. Furthermore, we hope the final metric can also reflect the degree of user fatigue in the interaction. Though \textit{Continuation} utterances are considered necessary in most cases, we think long dialogues should be avoided and better interaction models can be designed to shorten the length. To this end, we define \textbf{Fatigue Value} as the average thresholded length of tasks as shown in Eq. \eqref{eq:fatigue_value} -- if a task is longer than $\alpha$ turns ($\alpha$ is a preset parameter), we count its fatigue value as $\#(UTT_{\in T})-\alpha+1$ otherwise as 1. Then we define \textbf{Efficiency Rate} as shown in Eq. \eqref{eq:efficiency_rate}, which means the less reformulation or the shorter dialogue in each task, the more efficient we think a session is.

\begin{align}
\mathit{ReformRate}=&\frac{\#(UTT_{reform \in \mathbb{S}})}{\#(UTT_{\in \mathbb{S}})}\label{eq:reform_rate}
\end{align}
\begin{align}
FatigueValue=&\frac{\sum_{T\in \mathbb{S}}max(0, \#(UTT_{\in T})-\alpha)+1}{\#(TASK_{\in \mathbb{S}})}\label{eq:fatigue_value}
\end{align}
\begin{align}
\mathit{EfficiencyRate}=&\frac{1-\mathit{ReformRate}}{FatigueValue}\label{eq:efficiency_rate}
\end{align}

Lastly, we can define a unified \textbf{User Engagement Score} representing the overall user experience of a session. Here we define it as a plain arithmetic mean of both Success Rate and Efficiency Rate (Eq. \eqref{eq:ue_score}), but it can be extended to more sophisticated forms to fit specific cases and applications.

\vspace{-1ex}
\begin{align}
UE_{_{SCORE}}=&\frac{\mathit{SuccessRate}+\mathit{EfficiencyRate}}{2}\label{eq:ue_score}
\end{align}

Overall, this classification and corresponding metrics are conducted at the utterance level, which is easy-to-run for real-time systems. Previous methods mostly focus on task-level evaluations, however this can only be done at the end of each task and does not fit the needs of real-time systems. Additionally, they have to leverage an additional module to detect task boundaries. Furthermore, since the proposed user engagement status can indicate a positive/negative experience explicitly, the corresponding metrics are highly explainable and instructive for troubleshooting potential system problems. 



\vspace{-1ex}
\subsection{Datasets}
Since there does not exist dataset available for our study, we collect data from four intelligent assistants -- \textbf{DSTC2, DSTC3, Yahoo Captain (YCap), Google Home (GHome)} -- and annotate them\footnote{Code, annotated datasets as well as the annotation manual will be released for reproducing the experiment results.}. 

All dialogues take place between a human and a real system (called Human-Machine dialogues~\cite{serban2015survey}) 
, which fit our goal of evaluating real intelligent assistants. Moreover, they cover various task types, modalities and scenarios, based on which we can examine the performance and robustness of the proposed metrics and prediction models on a variety of cases.

\begin{table*}[!htbp]
  \begin{adjustwidth}{-0.25cm}{}
  \centering
  \fontsize{8}{8}\selectfont
  \renewcommand{\arraystretch}{1.0}
  \caption{Statistics of four annotated dialogue datasets}
  \vspace*{-2ex}
  \begin{tabular}{lC{0.9cm}C{1.0cm}C{1.5cm}rrrrrrccccc}
    \hline
    \hline
    \textbf{Dataset}
    & \textbf{\#(task)}
    & \textbf{\#(utt) per task}
    & \textbf{\#(word) per user\_utt}
    & \textbf{\textit{\#}(All)}
    & \multicolumn{1}{c}{\textbf{C\%}}
    & \multicolumn{1}{c}{\textbf{R\%}}
    & \multicolumn{1}{c}{\textbf{F\%}}
    & \multicolumn{1}{c}{\textbf{A\%}}
    & \multicolumn{1}{r}{\textbf{Success\%}}
    & \multicolumn{1}{r}{\textbf{Effic\%}}
    & \multicolumn{1}{r}{\textbf{Reform\%}}
    & \multicolumn{1}{r}{\textbf{Fatigue}}
    & \multicolumn{1}{r}{\textbf{$UE_{SCORE}$}}
    \\
    \hline
    \textbf{DSTC2} 
    & 2,825 & 4.36 & 3.87 & 5,700 & 28.6\% & 21.9\% & 47.1\% & 2.5\%
    & 93.8\% & 41.9\% & 17.0\% & 3.33 & 0.679
    \\
    \textbf{DSTC3} 
    & 3,020 & 4.64 & 4.00 & 5,856 & 28.1\% & 20.4\% & 48.0\% & 3.6\%
    & 90.1\% & 45.1\% & 14.6\% & 4.01 & 0.676
    \\  
    \textbf{YCap} 
    & 2,733 & 2.37 & 4.49 & 3,530 & 7.6\% & 14.9\% & 70.8\% & 6.6\%
    & 91.8\% & 78.7\% & 12.4\% & 1.35 & 0.853
    \\  
    \textbf{GHome}
    & 4,561 & 2.98 & 4.17 & 5,241 & 2.3\% & 10.6\% & 75.7\% & 11.4\%
    & 87.4\% & 73.3\% & 8.3\% & 1.80 & 0.804\\
    \hline
  \end{tabular} 
  \vspace*{-1.5ex}
\label{tab:dataset-characteristics-comparison}
  \end{adjustwidth}
\end{table*}

\textbf{DSTC2}~\cite{henderson2014second} and \textbf{DSTC3}~\cite{henderson2014third} are task-specific datasets, in which users call the system to inquire restaurant or tourist information. \textbf{YCap} is an SMS-based family assistant developed by Yahoo!. It supports functions like setting a reminder for family members, maintaining and sharing shopping list etc. \textbf{GHome} is collected from real users of Google Home, an intelligent home device powered by Google Assistant and responding to voice control with multiple functions. The \textbf{GHome} dataset is the most complicated among the four datasets. It not only covers a broad range of tasks including reminder, timer, search, in-house device control etc., but also supports open-domain chitchat. For \textbf{YCap} and \textbf{GHome}, as all the conversations are concatenated in a log file, we split dialogues by checking if the interval between two utterances is more than 10 minutes.
Then we randomly select 1,000 anonymized dialogues from each dataset for annotation. We ask professional annotators to judge the engagement status of each user utterance. The first pass of annotation is done by two annotators independently and the conflicts are resolved by the third annotator. The inter-annotator agreement achieves a kappa of 0.790, indicating the proposed scheme is understandable and easy-to-annotate. Table \ref{tab:dataset-characteristics-comparison} shows the statistics of each dataset. The dataset contains 4,000 annotated dialogues as well as 28,564 utterances. Here we highlight several observations: 
\begin{enumerate}
	\item The distributions of \textbf{DSTC2} and \textbf{DSTC3} are pretty similar. In fact, though DSTC3 offers more functions, the content and format of dialogues in two systems are not very different.
    \item By checking the average number of utterances (\#(utt) per task), dialogues of the text-based system (\textbf{YCap}) are averagely shorter than the ones of spoken systems. Also, since \textbf{YCap} takes the user typed input directly, though the data is intact from the error-prone ASR (Automatic Speech Recognition), it suffers from the typo errors of user inputs.
    \item \textit{Continuation} accounts for a large part in \textbf{DSTC2} and \textbf{DSTC3}. This is because, in order to inquire restaurants of interest, users have to interact with the system for many turns. But in \textbf{YCap} and \textbf{GHome}, the user requests are generally easier and most of them can be solved in one turn, such as ``set up a reminder at 8pm'' or ``turn on the light''.
    \item Utterances of \textit{Fulfillment} and \textit{Continuation} take the major part across all four datasets. By summing up these two types, we can see a basic success rate of each system at the utterance level (75.6\%:75.9\%:78.4\%:78.0\%). We also see that user \textit{Reformulation} and \textit{Abandonment} are common phenomenons across all datasets.
    \item \textit{Abandonment} on task-specific systems is notably fewer than on more complicated systems such as \textbf{GHome}, which can be attributed to the fact that tasks in \textbf{GHome} are more diverse and difficult.
    \item Overall, the class distribution is very skewed, and models may severely suffer from the data scarcity on minor classes.
\end{enumerate}

\subsection{Case study of User Engagement Metrics}
\begin{figure*}[h!]
    \centering
    \includegraphics[width=0.95\textwidth]{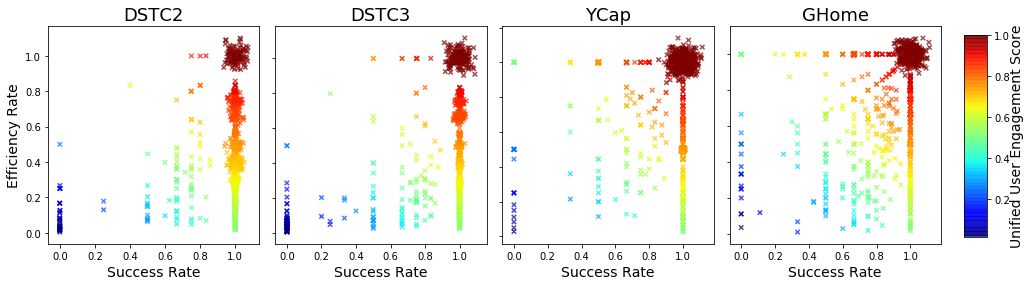}
    \caption{A 2-D Scatter plot of proposed user engagement metrics on four datasets. A jitter is added to show the size of clusters.}
    \label{fig:metric}
\end{figure*}

We compute the user engagement scores of each dataset as shown in Table \ref{tab:dataset-characteristics-comparison}. We also visualize the distribution of session scores in a 2-D scatter plot in Fig. \ref{fig:metric}. We set $\alpha$ to 2 for all datasets to discount tasks longer than 3 turns. From the table we can see that \textbf{YCap} and \textbf{GHome} perform overall better than the other two according to our metrics. All four assistants are able to achieve a satisfactory success rate, but \textbf{DSTC2} and \textbf{DSTC3} perform badly on efficiency. Specifically, among all the successful sessions ($SuccessRate=1.0$), the ratio of tasks whose efficiency is less than 0.5 is more than 50\% , but in \textbf{YCap} and \textbf{GHome} the percentage is less 20\%. Since the system used in DSTC datasets is considerably dated, we think the high \textit{Reformulation Rate} can be attributed to the poor ASR quality. What's more, we can also use the metric to quickly identify problematic dialogues, i.e. the ones have low engagement scores. There are 35/63/9/13 sessions whose overall score is less than 0.2. By manually examining those sessions, we find the most prominent issues in \textbf{DSTC2} and \textbf{DSTC3} are poor ASR and language understanding ability. A user may repeat 5 times to make the system understand what the request is about. \textbf{YCap} only takes user commands matching particular templates and oftentimes users reform their request several times to make it accepted. In \textbf{GHome}, problems are more diverse since it supports various functions and users can ask open-domain questions to which the system cannot handle well yet.

The goal here is to demonstrate how metrics based on the proposed user engagement status could be used to evaluate system performance and troubleshoot failures, and these metrics can be easily adopted for online A/B testing.

\section{Automatic Prediction of User Engagement Status}
Now we have defined a series of user engagement metrics for intelligent assistants, the next step is to automate the prediction of user engagement status so that the proposed metrics can be used in large-scale and online applications.


We naturally formalize the problem of predicting user engagement status as a four-class classification problem at the utterance-level, and we will concentrate on exploiting appropriate machine learning models, ranges of context and feature settings to automatically classify user utterances. 

\vspace{-1ex}
\subsection{Model Setting}
We mainly examine two groups of models. The first group is classic classifiers, working together with hand-crafted features. We consider three models which are broadly used for text classification: Logistic Regression (\textbf{LR}), Support Vector Machine (\textbf{SVM}) and Random Forest (\textbf{RF}). The second group is convolutional neural networks (\textbf{CNN}), which learn continuous representations without manual feature engineering and allow us to leverage word vectors pretrained on a large corpus, with which a significant performance boost has been observed in various NLP studies. We use two variants of CNNs proposed by~\citeauthor{kim2014convolutional}\cite{kim2014convolutional}: \textbf{CNN.Rand} and \textbf{CNN.MultiCh} (multi-channel). We have also tested a group of models based on recurrent neural networks, however they cannot converge well (may be due to the size of datasets). Thus their scores are not reported and discussed.

\vspace{-1ex}
\subsection{Context Setting}
The user engagement status greatly depends on the response from the system as well as the corresponding feedback of user. Previous studies have demonstrated the effects of contextual information in facilitating identification~\cite{bangalore2008learning, kim2010classifying}. 
By comparing different settings of context, we are able to know which utterances are most effective for predicting user engagement. We denote five utterances in time order as follows: 
\begin{olditemize}
\setlength\itemsep{0.25em}
\item \textbf{$user\_utt_{-1}$}: previous user utterance,
\item \textbf{$sys\_utt_{-1}$}: previous system utterance,
\item \textbf{$user\_utt_{0}$}: current user utterance,
\item \textbf{$user\_utt_{+1}$}: next user utterance,
\item \textbf{$sys\_utt_{+1}$}: next system utterance.
\end{olditemize}
And we define five settings of context as follows, covering different range of utterances in the dialogue: 
\begin{itemize}
\setlength\itemsep{0.5em}
\item \textbf{CUR\_UTT=\{$user\_utt_{0}$}\}, 
\item \textbf{CUR=\{$user\_utt_{0}$, $sys\_utt_{+1}$}\}, 
\item \textbf{NEXT=\{$user\_utt_{0}$, $sys\_utt_{+1}$, $user\_utt_{+1}$\}}, 
\item \textbf{PREV=\{$user\_utt_{-1}$, $sys\_utt_{-1}$, $user\_utt_{0}$\}}, 
\item \textbf{ALL=\{$user\_utt_{-1}$, $sys\_utt_{-1}$, $user\_utt_{0}$, $sys\_utt_{+1}$, $user\_utt_{+1}$\}}.
\end{itemize}

\vspace{-1ex}
\subsection{Feature Setting}
We think the status of user engagement is system-independent and identifiable by analyzing the dialogue contents. Therefore we only use features that can be extracted from transcriptions and ignore the other types of system-specific outputs (e.g., dialogue state, ASR output). From each utterance, we define eight groups of features and use them to predict user engagement status. Besides, we notice that \textit{Reformulation} implies a high semantic similarity between two user requests, thus we also define a set of \textit{similarity features} for each feature group.

\vspace{-1.0ex}
\subsubsection{Basic Features} 
It includes three subgroups of features that indicate basic information in each utterance. The first two subgroups are about utterance length (\textbf{utt\_length}) and time ($\textbf{if\_dialogue\_start}$, $\textbf{if\_dialogue\_end}$, $\textbf{\#utt\_from\_end}$, $\textbf{\#utt\_to\_end}$, $\textbf{time\_percent}$). The third one is about basic user commands. We manually build a vocabulary for each dataset, containing the most common user words or phrases (e.g. ``remind", ``alarm", ``add item") and based on which we define another three basic features: $\textbf{command\_word}$ (represented as one-hot vectors), $\textbf{\#command\_word}$, $\textbf{command\_jaccard\_similarity}$ (a similarity feature between two adjacent user utterances).

\vspace{-1.0ex}
\subsubsection{Phrasal Features}
Phrases and entities usually play an important role in representing users' intents. 
We apply the Stanford CoreNLP toolkit to extract 1) noun phrases ($\textbf{noun\_phrase}$) and 2) entities ($\textbf{entity}$) from each utterance and represent them as one-hot vectors. We define three similarity features: 3) $\textbf{repetition}$: if any noun phrase or entity is repeated in two adjacent user utterances; 4) $\textbf{\#repetition}$: number of repeated noun phrases or entities; 5) $\textbf{ngram\_jaccard\_similarity}$: Jaccard similarity of noun phrases or entities between adjacent user utterances.

\vspace{-1.0ex}
\subsubsection{Syntactic Features}
The syntactic dependencies can help us understand the core components of utterances. 
From the dependency tree of each utterance, we can extract three types of syntactic features and represent them as one-hot vectors: 1) root word ($\textbf{root\_word}$), 2) topmost subject word ($\textbf{subject\_word}$) and 3) topmost object word ($\textbf{object\_word}$). For similarity we only check if there is any repetition of these words between two user utterances: 4) $\textbf{repeat\_root\_word}$, 5) $\textbf{repeat\_subject\_word}$ and 6) $\textbf{repeat\_object\_word}$.

\vspace{-1.0ex}
\subsubsection{N-grams Features}
The n-grams is considered one of the most robust features for text classification.
We extract 1-, 2-, and 3-grams and represent them as one-hot vectors weighted by TfIdf. The similarity features for n-grams include: 1) $\textbf{edit\_distance}$ (Levenshtein edit distance between two user utterances) and 2) $\textbf{jaccard\_similarity}$.

\vspace{-1.0ex}
\subsubsection{Topic Features}
We apply the Latent Dirichlet Allocation (LDA) to capture the topical information in utterances ($\textbf{lda\_feature}$). We train separate LDA model for each dataset and set its dimension to 50. We use the cosine similarity of LDA vectors between two user utterances~($\textbf{lda\_cosine}$) as its similarity feature. 

\vspace{-1.0ex}
\subsubsection{Distributed Representations}
Previous studies~\cite{collobert2011natural, devlin2018bert} have demonstrated the efficacy of transferring language knowledge learned from rich resources to new tasks. Since we have only a limited amount of dialogues for training, we would like to know if we could follow the same way to alleviate data shortage. Here we present three utterance representations capturing different characteristics of language. They are all pre-trained on large external corpus (Google News or BookCorpus dataset), which have been widely applied to various text classification tasks~\cite{hill2016learning, kiros2015skip, le2014distributed}:
\textbf{Word2Vec}~\cite{mikolov2013distributed} (averaging word vectors in the utterance, dimension=300), \textbf{Doc2Vec}~\cite{le2014distributed} (treating each utterance as a document, dimension=300) and \textbf{Skip-thought}~\cite{kiros2015skip} (using bi-skip model, dimension=2400).

\section{Results of Automatic Prediction}
We conduct thorough empirical experiments on four datasets to demonstrate the effects of different machine learning models, context ranges as well as feature designs.
Specifically, we train and evaluate all models on each dataset using 10-fold cross-validation: 10\% for validation (to check if a model has converged), 10\% for testing and the rest 80\% for training. In order to perform significance tests on the relatively small datasets, we repeat the cross-validation five times, yielding 50 random splits and corresponding results. Unless otherwise stated, we report unweighted macro-average scores of 50 experiments on the testset. We apply two-sided paired T-test to examine the significance of changes. We also utilize the Bonferroni correction for T-test \cite{salzberg1997comparing, pizarro2002multiple, armstrong2014use}, a very conservative but safe approach, to counteract the risk of using overlapping data partitions (potentially leading to higher chance of making Type I errors).

\subsection{Comparison of Models}
We compare the performance of different models to get a general idea. We run experiments with the context range of $\textbf{ALL}$ to include as many features as we can. All three classic classifiers are trained with \textit{N-grams} features as well as similarity features. We report accuracy and F1-score, common metrics for classification tasks, of each model with optimal hyperparameters after a thorough grid search, in Table \ref{tab:model-comparison-context=all}. 

Two simple baseline models are compared here, outputting the major class in the training set (Majority) or a random class uniformly (Random). Both simple baselines work poorly, and the F1-score of Majority is even lower due to the very skewed class distribution. The primary models perform fairly well. The two CNN models, without any human-designed feature, outperform all the other models in the current setting. The benefit of adopting pre-trained word vector is slight but significant ($p_{value}$$<$0.01). 

It shows comparative performance among the three classic models. The SVM performs the best, but its advantage over LR is marginal ($p_{value}>$0.05). Thus for the rest of this study, we only present and discuss the results of Logistic Regression, taking the advantage of its interpretability on feature importance. Specifically, we use the Logistic Regression with the L1 regularization ($\lambda=1.0$), which performs robustly across different features and datasets.


\begin{table}[!htbp]
  \centering
  \fontsize{9}{9}\selectfont
  \renewcommand{\arraystretch}{1.0}
  \vspace{-2ex}
  \begin{center}
  \caption{The averaged performance of user engagement classification with different models on four datasets (context=\textbf{All}, with similarity features). The underline indicates the maximum value in each column.}
  \vspace{-2ex}
  \begin{tabular}{lcc}
    \hline
    \hline
    \textbf{Model}
    & \textbf{Accuracy}
    & \textbf{F1-score}
    \\
    \hline
    Majority & 0.6020 & 0.1858
    \\ 
    Random & 0.2503 & 0.2029
    \\
    \hline
    SVM & 0.8410 & 0.6440
    \\
    LR & 0.8398 & 0.6413
    \\
    RF & \underline{0.8415} & 0.6192
     \\
    \hline
	CNN.Rand & 0.8287 & 0.6549
    \\
	CNN.MultiCh & 0.8367 & \underline{0.6674}
    \\
    \hline
  \end{tabular}
  \vspace{-4ex}
  \label{tab:model-comparison-context=all}
  \end{center}
\end{table}

\subsection{Comparison of Context Settings}
In this subsection, we mainly investigate which utterances are most important for detecting user engagement status. In order to predict the class of current user utterance, the next system response and the corresponding user feedback are supposed to play important roles. Also, as stated by~\cite{kim2010classifying}, additional contextual information is helpful in classifying dialogue acts, thus we would like to see if the same observation applies to our task. We list the performance comparison with five context settings in Table \ref{tab:context-setting-comparison}. Note that, since there is no similarity feature for \textbf{CUR\_UTT} and \textbf{CUR}, in order to fairly compare the contributions of different contexts, we exclude all similarity features for these experiments.

Firstly, we see that, the score difference is consistent across different context settings, indicating that the context is a significant factor in engagement status prediction. \textbf{CUR\_UTT} performs the worst among the five settings, since it includes only the content of the current user utterance and it provides very limited information. As for \textbf{CUR}, with one system utterance, the performance is remarkably better than the \textbf{CUR\_UTT}. Furthermore, with the evident feedback from user ($user\_utt_{+1}$), \textbf{NEXT} performs generally the best among all context settings. This result conveys a clear message that, the following utterances from both system and user are critical in determining whether the next system response is relevant or not and whether the user is satisfied or not.

As for \textbf{PREV} and \textbf{ALL}, which include the historical information of user requests, the performances are generally no better than the \textbf{CUR\_UTT} and \textbf{NEXT} respectively. But the negative effects on distributed representations and models are much less than on the rest features, especially for CNN. We speculate this is because most user requests can be satisfied within a few turns and do not require much historical information, thus the features from previous utterances rarely take effect and even become detrimental. According to the error analysis, we do observe several examples to which knowing the long period of history is important. Determining in what cases the historical information is in effect would be beneficial.

\begin{table}[!htbp]
  \centering
  \fontsize{7.5}{8}\selectfont
  \renewcommand{\arraystretch}{1.0}
  \caption{The performance (F1-score) comparison of user engagement classification with different context settings (without similarity features). $\dagger/\ddagger$ indicates a statistical significant difference at p$<$0.05/p$<$0.01 between \textbf{CUR\_UTT} and \textbf{PREV} or between \textbf{NEXT} and \textbf{ALL}. The underline indicates the best score in each column and the bold font indicates the best in each row.}
  \begin{tabular}{L{1.4cm}|C{1.3cm}C{0.8cm}C{0.8cm}C{1.0cm}C{1.0cm}}
    \hline
    \hline
    \textbf{Model}
    & \textbf{CUR$\_$UTT}
    & \textbf{CUR}
    & \textbf{NEXT}
    & \textbf{PREV}
    & \textbf{ALL}
    \\    
    \hline    
    Basic & 0.3425 & 0.3503 & 0.3836 & 0.3501$\dagger\ddagger$ & \textbf{0.3963}$\dagger\ddagger$\\  
    Phrasal & 0.3679 & 0.5521 & \textbf{0.5913} & 0.3709$\quad$ & 0.5661$\dagger\ddagger$\\  
    Syntactic & 0.3485 & 0.5530 & \textbf{0.6078}  & 0.3671$\dagger\ddagger$ & 0.5867$\dagger\ddagger$\\  
    N-grams & 0.3839 & 0.5694 & \textbf{0.6113} & 0.3788$\quad$ & 0.5984$\dagger\ddagger$\\  
    Topic Model & 0.2982 & 0.5255 & 0.5803 & 0.3464$\dagger\ddagger$ & \textbf{0.5829}$\quad$\\
    Word2Vec & 0.3704 & 0.5723 & \textbf{0.6162} & 0.3827$\dagger\ddagger$ & 0.6032$\dagger\ddagger$\\  
    Doc2Vec & 0.3427 & 0.5379 & \textbf{0.5858} & 0.3722$\dagger\ddagger$ & 0.5740$\dagger\ddagger$\\  
    Skip-thought & 0.3648 & 0.5545 & \textbf{0.6063} & 0.3692$\quad$ & 0.6008$\dagger\;\,$\\ \hline

    CNN.Rand & \underline{0.4252} & \underline{0.5862} & \textbf{0.6647}  & 0.4153$\quad$ & 0.6549$\dagger\;\,$\\ 
    CNN.MultiCh & 0.4207 & 0.5829 & \underline{\textbf{0.6685}} & \underline{0.4288}$\quad$ & \underline{0.6674}$\quad$\\  
    \hline
    \hline
  \end{tabular} 

  \label{tab:context-setting-comparison}
\end{table}

\subsection{Effects of Similarity Features}
Based on the comparison of context settings, here we focus on analyzing the models with \textbf{NEXT} setting. We show the performances of Logistic Regression with and without similarity features in Table \ref{tab:comparison_similarity}. By adding similarity features, which are just one or two additional features, the scores on different feature groups increase significantly. The similarity features are devised to facilitate detecting the reformulated utterances, and we observe that the average improvement on the \textit{Reformulation} (8.06\%) is much more salient than other three classes (2.99\%, 1.74\% and 0.51\%). Feature importance analysis based on one-way ANOVA shows that similarities on \textit{N-gram}, \textit{LDA} and \textit{Phrasal} features are most significant, which is consistent with the improvement in Table~\ref{tab:comparison_similarity}.

\begin{table}[!htbp]
  \centering
  \fontsize{9}{9}\selectfont
  \renewcommand{\arraystretch}{1.0}
  \caption{The performance comparison of user engagement classification without and with similarity features (context=\textbf{NEXT}). $\dagger/\ddagger$ indicates a statistical significant change at p$<$0.05/p$<$0.01 between results with and without similarity features. The bold font/underline indicates the maximum value in the respective row/column.}
\label{tab:comparison_similarity}
\vspace{-2ex}
  \begin{tabular}{l|cc}
    \hline\hline
    \textbf{Model}
    & \textbf{w\textbackslash o Similarity}
    & \textbf{w\textbackslash  Similarity}
    \\    
    \hline    
    Basic & 0.3836 & \textbf{0.4105} (+2.69\%)$\quad$\\  
    Phrasal & 0.5913 & \textbf{0.6316} (+4.03\%)$\dagger\ddagger$\\  
    Syntactic & 0.6078 & \textbf{0.6280}	(+2.02\%)$\dagger\ddagger$\\  
    N-grams & 0.6113 & \underline{\textbf{0.6573}} (+4.60\%)$\dagger\ddagger$\\  
    Topic Model & 0.5803 & \textbf{0.6346} (+5.43\%)$\dagger\ddagger$\\
    Word2Vec & \underline{0.6162} & \textbf{0.6521} (+3.59\%)$\dagger\ddagger$ \\  
    Doc2Vec & 0.5858 & \textbf{0.5968} (+1.10\%)$\quad$\\  
    Skip-thought & 0.6063 & \textbf{0.6216} (+1.53\%)$\quad$\\  
    \hline\hline
  \end{tabular} 
  \vspace{-3ex}
\end{table}

\subsection{Analysis on Feature Groups}
\label{subsec:feature_group}
\begin{table*}[!htbp]
  \centering
  \fontsize{8}{8}\selectfont
  \renewcommand{\arraystretch}{1.0}
  \caption{The F1-score of user engagement classification with different features (context=\textbf{NEXT}, with similarity features). The right part presents F1-score of best models on each dataset. The underline indicates the best score in each column. The bold indicates the better score between models with and without feature selection. $\dagger/\ddagger$ indicates a statistical significant difference at p$<$0.05/p$<$0.01.}
  \vspace{-2ex}
  \begin{tabular}{l|c|cc|C{1.2cm}C{1.2cm}C{1.2cm}C{1.2cm}}
    \hline\hline  
    \multicolumn{1}{c|}{\multirow{2}{*}{\textbf{Model}}}
	& \textit{\textbf{w\textbackslash o FeatSelect}}
    & \multicolumn{2}{c|}{\textit{\textbf{w\textbackslash  \textit{FeatSelect}}}}
    & \multirow{2}{*}{\textbf{DSTC2}}
    & \multirow{2}{*}{\textbf{DSTC3}}
    & \multirow{2}{*}{\textbf{YCap}}
    & \multirow{2}{*}{\textbf{GHome}}
    \\    
    \cline{3-4} 
    & \textbf{w\textbackslash \textit{Sim}} 
    & \textbf{w\textbackslash o \textit{Sim}} & \textbf{w\textbackslash \textit{Sim}} & & & &
    \\
    \hline    
    
    (a) Basic
     & 0.4105 & - & 0.4105$\quad$
     & 0.5411 & 0.5044 & 0.3079 & 0.2886
    \\
    (b) Phrasal
     & 0.6316 & - & \textbf{0.6318}$\quad$
     & 0.6470 & 0.6703 & 0.6593 & 0.5508
    \\
    (c) Syntactic 
    & 0.6280 & - & \textbf{0.6402}$\dagger\ddagger$
     & 0.6567 & 0.6469 & 0.7005 & 0.5566
    \\
	(d) N-grams 
	 & 0.6573 & - & \textbf{0.6770}$\dagger\ddagger$
     & 0.7078 & 0.6905 & 0.6851 & 0.6248
    \\ 
	(e) Topic model 
     & 0.6346 & - & \textbf{0.6358}$\quad$
     & 0.6774 & 0.6384 & 0.6397 & 0.5877
    \\
	(f) Word2Vec 
     & 0.6521 & - & \textbf{0.6523}$\quad$
     & 0.6919 & 0.6919 & 0.6209 & 0.6043
    \\
	(g) Doc2Vec 
     & 0.5968 & - & \textbf{0.5969}$\quad$
     & 0.6325 & 0.6335 & 0.5730 & 0.5486
    \\
	(h) Skip-thought 
     & 0.6216 & - & \textbf{0.6216}$\quad$
     & 0.6654 & 0.6414 & 0.6020 & 0.5775
    \\
    \hline  
    
    (i) (a) + (b) + (c) + (d)  & 0.6694 & 0.6511 & \textbf{0.7085}$\dagger\ddagger$
    & 0.7360 & 0.7151 & 0.7218 & 0.6613
    \\  
    \multicolumn{1}{l|}{(j)~~\hspace{0.5cm}+ Topic Model} & 0.6720 & 0.6617
    & \underline{\textbf{0.7152}}$\dagger\ddagger$
    & 0.7438 & 0.7161 & \underline{0.7314} & \underline{0.6699}
    \\
    \multicolumn{1}{l|}{(k)~\hspace{0.5cm}+ Word2Vec}
    & 0.6790 
    & 0.6617 & \textbf{0.7135}$\dagger\ddagger$
    & \underline{0.7514} & 0.7194 & 0.7180 & 0.6651
    \\  
    \multicolumn{1}{l|}{(l)~~\hspace{0.5cm}+ Doc2Vec}
    & 0.6713 
    & 0.6631 & \textbf{0.7100}$\dagger\ddagger$
    & 0.7390 & 0.7149 & 0.7269 & 0.6592
    \\  
    \multicolumn{1}{l|}{(m)\hspace{0.5cm}+ Skip-thought} & 0.6747 & \underline{0.6666}
    & \textbf{0.7124}$\dagger\ddagger$
    & 0.7412 & 0.7181 & 0.7209 & 0.6696
    \\  
    \hline
	(n) All & \underline{0.6825} 
    & 0.6589 &  \textbf{0.7140}$\dagger\ddagger$
    & 0.7490 & \underline{0.7213} & 0.7202 & 0.6655
    \\
    \hline
	(o) CNN.Rand & 0.6647 &  -
     & - & 0.6798 & 0.6669 & 0.6943 & 0.6176
    \\
	(p) CNN.MultiCh & 0.6685 &  -
     & - & 0.6880 & 0.6612 & 0.7054 & 0.6196
    \\
    \hline\hline
  \end{tabular} 
  \label{tab:comparison-feature-groups}
\end{table*}

Furthermore, we apply another two techniques to obtain the optimal model performance: feature combination and feature selection. On one hand, the first four feature groups are discrete and capture various local linguistic information, but the latter four groups give continuous representations with regard to the whole utterance. Thus we consider combining these two sets of features and expect further improvement with the advantages of both. On the other hand, feature selection has been proved helpful in reducing noisy features. Here we apply \textit{Chi-square statistic} to discrete feature groups and \textit{Principal Component Analysis (PCA)} to continuous feature groups. We report the best performance of each setting after a grid search (feature numbers in the exponential power of 2) in Table \ref{tab:comparison-feature-groups}. 

Overall, we observe that most models with feature selection outperform the original ones significantly. The feature selection works more significantly on groups having a large number of features such as \textit{N-grams}, \textit{Syntactic} and combined feature groups, indicating that only a small proportion of discrete features is actually in effect. Also the performances on combined feature groups (row \textbf{i} to \textbf{n}) are much better than on any of individual groups. But we observe that the continuous representations (\textbf{j-n}) contribute marginally on the top of the combined discrete features (\textbf{i}). 

With the help of these two improvements on features, the Logistic Regression outruns the previous best model CNN by a large margin. But if we exclude the similarity features (3rd column), we find that CNN still works on a par with the best LR models.  
Since the CNN is only trained to distinguish classes and it does not take any explicit input about similarity, the best LR models with similarity features beat CNN soundly. In order to let the CNN be aware of the user reformulation, we think it might be helpful to leverage a submodule for similarity calculation: train the submodule separately in a way like paraphrase identification~\cite{yin2015convolutional}, and take the similarity vector as additional input for classification. 

We observe that deep learning approaches pre-trained with external resources, no matter representations (\textbf{f-h} and \textbf{k-m}) or models (\textbf{p}), do not show superior performance over models trained locally. This may be attributed to two reasons. One is that external representations are trained on the Google News or BookCorpus, of which text genre is formal and far different from dialogues. Another reason is that all these methods represent each utterance as a whole, but many words inside are not useful for understanding real user intent. Conversely, discrete feature groups are good at capturing key information. We conjecture that models trained on more local texts would lead to better results~\cite{lowe2017towards}. It also leaves a question open about whether it is beneficial to incorporate discrete features into CNNs.

Table \ref{tab:comparison-feature-groups} also presents detailed scores on each dataset after feature selection. One trend emerging among most LR results is that, the performances decrease gradually from \textbf{DSTC2} to \textbf{GHome}, implying the difficulty of each dataset. \textit{LR+Basic} works well on \textbf{DSTC2} and \textbf{DSTC3} but poorly on the other two datasets. As we know, the \textit{command\_word} in \textit{Basic} covers the most common user commands, and therefore it performs adequately in simple dialogues. But in more complicated cases, general words or linguistic components from both user and system sides become necessary, such as confirmations (ok, sure, yeah, etc.), success and failure signals (
discard, sorry, don't understand, etc.), function-related words, and they are captured in different feature groups.

\section{Discussion}

\subsection{Analysis on Failure Cases}
To understand better what major shortcomings our current models suffer from, we manually examine 50 random wrongly-predicted examples from \textbf{GHome} dataset and try to understand the reasons behind. Among the 50 failures, 22 examples of \textit{Reformulation}, 21 of \textit{Abandonment}, 4 of \textit{Fulfillment} and 3 of \textit{Continuation} are misclassified by the model. The highly skewed class distribution might be one major reason. The model is trained with very few examples of \textit{Reformulation} and \textit{Abandonment}, therefore it is more prone to make mistakes on them. 

We also notice some issues that are general to all dialogue related tasks, which might be difficult to overcome with the NLP techniques used in this study. (1) A common error (16 failures) we note is that the model cannot distinguish whether a system response is relevant to a user's request or not. Our models can only determine the relevance by feature matching instead of understanding the actual semantics, therefore it often fails to predict, particularly when the user request is long or task-general. (2) We find 15 examples that require taking into account the contextual and historical information. For example, a user asks the Google Home to ``Turn the Christmas tree off'' and ``Turn it on'', our model does not recognize ``it'' refers to the previous ``Christmas tree'' and makes mistake. Though we have tried to incorporate the content of the previous turn as a short history, it is still difficult to learn such coreference and dependency. Another long-dependency case is, the system confirms a similar question after a few turns, which should be considered as \textit{Reformulation}, but this can be hardly addressed by our current models. (3) The third common mistake is more specific to \textit{Reformulation}, which occurs in 9 examples. On one hand, a user may paraphrase an utterance in a different way to help the system understand, such as 
from ``I want the stair lights'' to ``turn on the stair lights''. On the other hand, a user can also issue two apparently similar but different requests, say ``how skinny is my husband'' and ``how old is my husband''. Although we utilize a range of similarity features, oftentimes the model still cannot identify the semantic resemblance or difference between them. A more powerful semantic encoder~\cite{devlin2018bert} might be helpful in this case.

Lastly, we note that in some cases user utterances are not recognizable due to auto speech recognition (ASR) errors. For example, a user says ``volume app'' and then corrects to ``volume up''. The mistranscribed data can badly affect the understanding of the text, especially for the \textbf{\textit{Reformulation}} detection as our similarity features are mostly based on words. Incorporating more information from the upstream ASR outputs might be a good way to alleviate this problem. 


\subsection{Analysis on Feature Importance}
With the help of the interpretability of Logistic Regression (the model weight of \textbf{LR} on each feature greatly represents its importance for prediction), we are able to know what features are most important. Here we only analyze our best model (model \textit{j} in Table \ref{tab:comparison-feature-groups}), which contains 256 discrete features (the first four groups) after the feature selection and 32 LDA features after the dimension reduction, with the \textbf{NEXT} context setting ($user\_utt_{0}$, $sys\_utt_{+1}$, $user\_utt_{+1}$).

\begin{figure}[h!]
    \centering
    \includegraphics[width=0.35\textwidth]{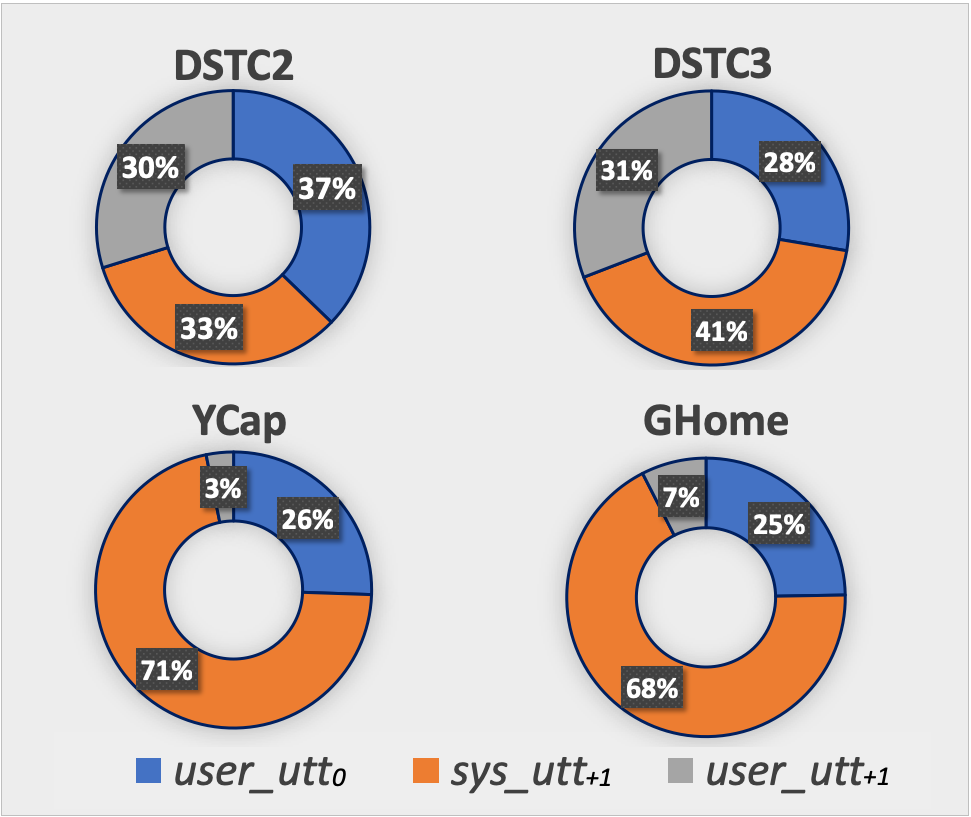}
    \vspace{-2ex}
    \caption{Pie charts showing the distribution of most important features in four datasets.}
    \vspace{-3ex}
    \label{fig:top100_distribution_figure}
\end{figure}

We first examine the distribution of the top 100 features from different utterances across different datasets, as shown in Figure \ref{fig:top100_distribution_figure}. Generally the next system utterance $sys\_utt_{+1}$ contributes the most features, ranging from 33\% (\textbf{DSTC2}) to 71\% (\textbf{YCap}), which is probably because that the system response is typically template-based, from which it is easy for the model to determine which user request/task a data point is about, then combining with other features to make predictions. But for \textbf{YCap} and  \textbf{GHome} almost 70\% of features come from the system response and the next user utterances do not contribute very much. This might be because many requests are fulfilled with only one turn of interaction and usually they are independent of each other, therefore the model does not need to rely on the next user utterance to predict. We manually check the features from next user utterances and find that most features are about similarities and confirmation words (yes, today, no, etc.), which are useful for detecting \textit{\textbf{Reformulation}} and \textit{\textbf{Continuation}}. As for the two DSTC datasets, we find that words such as 'address', 'phone', 'area', 'bye', which often imply a successful transit to another user request, are ranked as top features, indicating a unique characteristic of DSTC datasets that one dialogue usually consists of multiple user requests.

\begin{table}[!htbp]
    \centering
    \vspace{-2ex}
    \fontsize{8}{8}\selectfont
    \caption{Top features ranked by model weights (\textbf{\textit{on GHome}}).}
    \label{tab:ghome_feature_example}
  	\vspace{-2ex}
    \begin{tabular}{rccc}
        \toprule\hline
		\# & Context & Feature Type & Name  \\\hline
        1 & $user\_utt_{0}$ & Topic & topic\_1  \\\hline
        2 & $sys\_utt_{+1}$ & N-gram & sorry  \\\hline
        3 & $sys\_utt_{+1}$ & Phrasal & ok  \\\hline
        4 & $sys\_utt_{+1}$ &  N-gram & don't  \\\hline
        5 & $global$ &  Basic & \textbf{if\_dialogue\_end}  \\\hline
        6 & $user\_utt_{+1}$ &  N-gram & \textbf{jaccard\_similarity}  \\\hline
        7 & $sys\_utt_{+1}$ & Phrasal & playing  \\\hline
        8 & $sys\_utt_{+1}$ & Phrasal & turning  \\\hline
        9 & $sys\_utt_{+1}$ & Phrasal & seconds  \\\hline
        10 & $sys\_utt_{+1}$ & Phrasal & minutes  \\\hline
   	   \bottomrule
    \end{tabular}
    \vspace{-2ex}
\end{table}

We also analyze the distribution of different feature types. Among the top 100 features in the \textbf{GHome}, we find 19 n-grams features, 25 phrasal features, 30 syntactic features, 17 LDA topic features and the rest are basic and similarity features. The relatively even distribution indicates that different feature sets are complementary to each other to obtain the best model. Table \ref{tab:ghome_feature_example} list the top 10 features of \textbf{GHome}. Except for the first topic feature, all the other features appear frequently in the dataset and each of them can be seen as a significant signal for a specific utterance type. For example, Both ``sorry'' and ``don't'' in the system utterances are strong indicators of a system error (``Sorry, I don't know how to help with that.''). The last four phrasal features are confirmative replies indicating the requested tasks are executed (i.e. playing music, turning light on/off and setting a timer). Overall, the high-frequency task-specific words play an important role in classifying utterances, thus we believe that a complete function vocabulary would be of great help.

\section{Conclusion and Future Work}
In a preliminary effort to solve the challenging problem of online evaluation for large-scale intelligent assistants, 
we provide a practicable solution, by converting the problem into a more tractable classification task and automating it with various machine learning methods. We admit there is still a long way to go for our model to work well in real environments. Also, more research is in urgent need to bridge the gap between utterance-level user engagement status and task- or session-level user experience. 
Thus, for the future work, we will first apply online A/B testing to validate whether any of proposed utterance-level user engagement status and metrics correlates well with the real long-term success. Furthermore, we will conduct offline user studies to rethink the basic level for user-oriented evaluation since neither simple utterance-level nor vague task-level is sufficient in practice. We believe with insights from these studies, we can understand the user experience in intelligent assistants better and subsequently propose better evaluation methods.

 

\bibliographystyle{ACM-Reference-Format}
\bibliography{umap2020}

\clearpage

\appendix

\section{Data Examples}
\label{appd:data_examples}
Three examples of annotated dialogue from \textbf{DSTC3}, \textbf{YCap} and \textbf{GHome} as shown in Table \ref{tab:dstc3_example}, Table \ref{tab:ycap_example} and Table \ref{tab:ghome_example} respectively.

\begin{table}[!htbp]
    \centering
    \footnotesize
    \caption{An annotated dialogue from the \textbf{\textit{DSTC3}} dataset.}
    \label{tab:dstc3_example}
  	\vspace{-2ex}
    \begin{tabular}{rL{6.2cm}|C{0.5cm}}
        \toprule\hline
        \textbf{Bot:} & Thank you for calling the Cambridge Information system. Your call will be recorded for research purposes. You may ask for information about a place to eat, such as a restaurant, a pub, or a cafe. How may I help you? &  \\\cdashline{2-2}
		\textbf{User:} & Looking for a continental restaurant & R\\ \hline
		\textbf{Bot:} & Are you looking for a pub or a restaurant? &  \\\cdashline{2-2}
		\textbf{User:} & Continental restaurant & C\\ \hline
		\textbf{Bot:} & What kind of price range are you looking for? For example, cheap, or free. &  \\\cdashline{2-2}
		\textbf{User:} & Expensive & R\\ \hline
		\textbf{Bot:} & What kind of price range are you looking for? For example, cheap. &  \\\cdashline{2-2}
		\textbf{User:} & Expensive & C\\ \hline
		\textbf{Bot:} & What part of town do you have in mind? For example, something in the Barnwell, Castle Hill, or Romsey area. &  \\\cdashline{2-2}
		\textbf{User:} & Doesnt matter & F\\ \hline
		\textbf{Bot:} & Alimentum, which serves continental food, is an outstanding restaurant. It serves french food and it's in the expensive price range. &  \\\cdashline{2-2}
		\textbf{User:} & Ah whats the phone number and area & R \\ \hline
		\textbf{Bot:} & The phone number of Alimentum is 01223413000. &  \\\cdashline{2-2}
		\textbf{User:} & And whats the area & F\\ \hline
		\textbf{Bot:} & Alimentum is in the Addenbrooke's part of town. &  \\\cdashline{2-2}
		\textbf{User:} & Thank you good bye &   \\ \hline
        
        \bottomrule
    \end{tabular}
\end{table}

\vspace{-2ex}
\begin{table}[!htbp]
    \centering
    \footnotesize
    \caption{An annotated dialogue from the \textbf{\textit{YCap}} dataset.}
    \label{tab:ycap_example}
  	\vspace{-2ex}
    \begin{tabular}{rL{6.2cm}|C{0.5cm}}
        \toprule\hline
		\textbf{User:} & @remind us to Book Tickets on 03/15/2017 at 8:50 am & R \\\cdashline{2-2}
        \textbf{Bot:} & Couldn't notify @remind. You can only notify family members. & \\\hline
        
		\textbf{User:} & remind us to Book Holi Tickets on 03/15/2017 at 8:50 am & F \\\cdashline{2-2}
        \textbf{Bot:} & Ok, I'll remind you and xxxxxx, 'Book Holi Tickets' on Wed, Mar 15 at 8:50 AM. & \\\hline
		
		\textbf{User:} & Edit Wednesday 11am. & A \\ \cdashline{2-2}
        \textbf{Bot:} & I'm learning. Can you clarify? &  \\\hline
		
		\textbf{User:} & Cancel reminder with lunch with xxxxxx on Thursday at 11am & C \\ \cdashline{2-2}
        \textbf{Bot:} & Did you mean to remove event 'Lunch with xxxxxx'? (yes/no) &  \\\hline
		
		\textbf{User:} & Yes & F \\ \cdashline{2-2}
        \textbf{Bot:} & Ok, I've removed the reminder: 'Lunch with xxxxxx'. &  
        
        \\\hline
        \bottomrule
    \end{tabular}
    \vspace{-4ex}
\end{table}

\begin{table}[htbp]
    \centering
    \footnotesize
    \caption{An annotated dialogue from the \textbf{\textit{GHome}} dataset.}
    \label{tab:ghome_example}
  	\vspace{-2ex}
    \begin{tabular}{rL{6.2cm}|C{0.5cm}}
        \toprule\hline
		\textbf{User:} & Turn off Aiden's room & A \\\cdashline{2-2}
        \textbf{Bot:} & Sorry, I'm not sure how to help with that. & \\\hline
		\textbf{User:} & Turn off my bedroom & F \\\cdashline{2-2}
        \textbf{Bot:} & You got it, turning the Bedroom off. & \\\hline
		\textbf{User:} & Turn off the big light & A \\\cdashline{2-2}
        \textbf{Bot:} & Sorry, I'm not sure how to help. & \\\hline
		\textbf{User:} & Play La Bamba & F \\\cdashline{2-2}
        \textbf{Bot:} & La Bamba by Ritchie Valens, sure. Playing on YouTube... & \\\hline
		\textbf{User:} & Turn on the garage door & F \\\cdashline{2-2}
        \textbf{Bot:} & You got it, turning the Garage Door on. & \\\hline
        \bottomrule
    \end{tabular}
\end{table}